\pdfoutput=1

\documentclass[11pt]{article}

\usepackage{ACL2023}

\usepackage{times}
\usepackage{latexsym}

\usepackage[T1]{fontenc}

\usepackage[utf8]{inputenc}

\usepackage{microtype}

\usepackage{inconsolata}

\usepackage{algorithm}
\usepackage{algorithmic}
\usepackage{graphicx}
\usepackage{caption}
\usepackage{subcaption}
\usepackage{diagbox}
\usepackage{expex}
\usepackage{dialogue}
\usepackage{microtype}
\usepackage{makecell}
\usepackage{booktabs}
\usepackage{todonotes}
\usepackage{adjustbox}

\newcommand{\oneS}{\ensuremath{{}^{\textstyle *}}}

\title{How About Kind of Generating Hedges using End-to-End Neural Models?}

\author{Alafate Abulimiti$^{1,2}$,
Chloé Clavel$^{3}$, 
Justine Cassell$^{1,4}$
 \\~\\
 $^1$  INRIA, Paris $^2$ ENS/PSL 
  \texttt{<alafate.abulimiti@inria.fr>} 
\\
$^3$ LTCI, Insitut Polytechnique de Paris, Telecom Paris 
{\tt <chloe.clavel@telecom-paris.fr>}\\
$^4$  Carnegie Mellon University 
\texttt{<justine@cs.cmu.edu>}
}

\begin{document}
\maketitle
\begin{abstract}
Hedging is a strategy for softening the impact of a statement in conversation. In reducing the strength of an expression, it may help to avoid embarrassment (more technically, ``face threat'') to one's listener. For this reason, it is often found in contexts of instruction, such as tutoring. In this work, we develop a model of hedge generation based on \textit{i)} fine-tuning state-of-the-art language models trained on human-human tutoring data, followed by \textit{ii)} reranking to select the candidate that best matches the expected hedging strategy within a candidate pool using a hedge classifier. We apply this method to a natural peer-tutoring corpus containing a significant number of disfluencies, repetitions, and repairs. The results show that generation in this noisy environment is feasible with reranking. By conducting an error analysis for both approaches, we reveal the challenges faced by systems attempting to accomplish both social and task-oriented goals in conversation.
\end{abstract}

\section{Introduction}

When people interact, they attend not just to the task at hand, but also to their relationship with their interlocutors \citep{tracy1990multiple}. One key aspect of the relationship that people attend to, while engaging in contexts as diverse as sales \citep{gremler2008rapport, planken2005managing}, education \citep{glazier2016building, murphy2012rapport} and healthcare \citep{dimatteo1979social, leach2005rapport}, is what is referred to as \textit{rapport}, a sense of harmony and mutual understanding between participants in a conversation \citep{spencer2005politeness,tickle1990nature}. Indeed, higher levels of rapport are correlated with better performance in each of these domains. \citet{zhao2014towards} describes rapport as built upon a base of mutual attentiveness, face management, and coordination. This base is built primarily by conversational strategies, or ways of speaking (including nonverbal and paraverbal behaviors) that manage rapport throughout a conversation. Key conversational strategies include self-disclosure, reference to shared experience, praise, and \textit{hedging} --- giving instructions or conveying information in an indirect manner when it might otherwise sound rude or overly demanding.

End-to-end large language models (LLM), of the kind that are increasingly popular and powerful, do a good job at carrying out the propositional or information-carrying aspects of conversation, and a relatively good job of maintaining the coherence of a conversation, but they are not as good at changing \textit{how} they say something as a function of a relationship with the human user, while humans are, for the most part, quite good at this. However, since saying things in a specific manner - for example, through a hedge - helps task performance, it is an important topic for dialogue systems. 

\begin{figure}
    \centering
    \includegraphics[width=0.6\linewidth, scale=0.2]{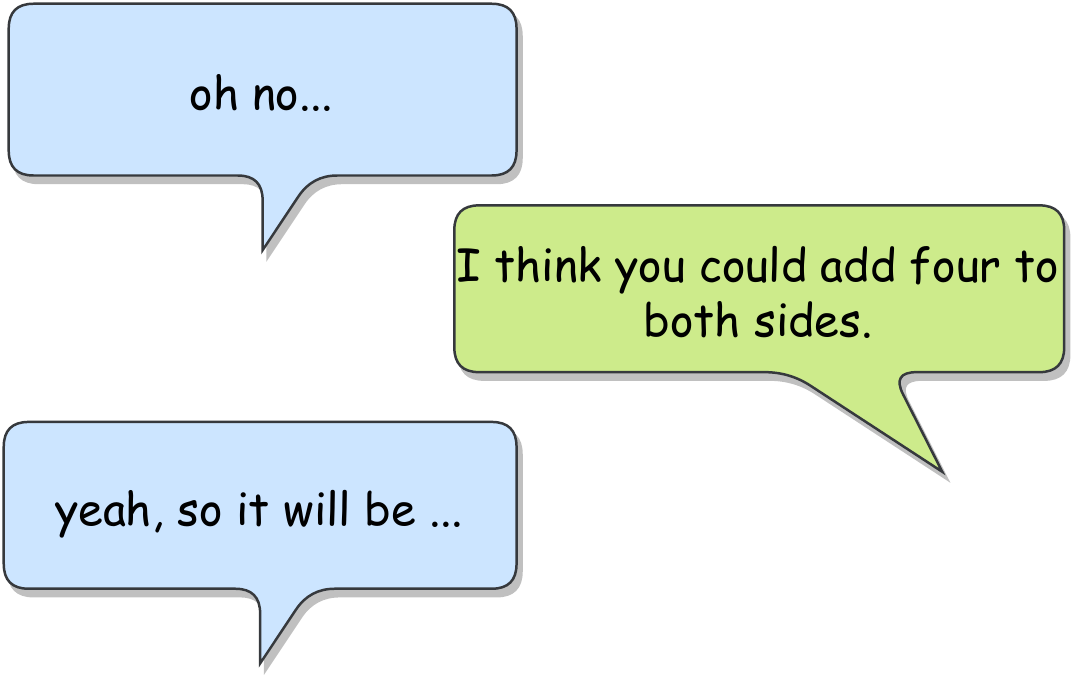}
    \caption{\footnotesize{Hedging in peer tutoring}}
    \label{fig:example}
\end{figure}

Linguists define hedges as a way of diminishing face threat (meaning the ``positive social value a person effectively claims for himself'' \citep{goffman1967} by attenuating the extent or impact of an expression \citep{brown1987a, fraser2010a}. Figure \ref{fig:example} shows a typical example of hedging in a peer tutoring setting, where the tutor uses two hedges (``I think'' and ``could'' rather than ``should'') to deliver a hint for the next step of solving an algebra equation.

Tutoring is one context in which hedges are found in abundance and where recognizing them might be important for intelligent tutoring systems, as attested by the number of computational approaches that attempt to do so (see section \ref{related_work}). Interestingly, even unskilled tutors use them. In fact, research on peer tutoring has shown that when rapport between a peer tutor and tutee is low, but the tutor is confident in his/her skills, that tutor tends to use more hedges, and this results in more problems attempted by the student and more problems successfully solved \citep{madaio2017impact}.

In this paper, then, we work towards the development of a generation module for a virtual peer tutor that, like real peer tutors, is able to choose the manner of delivering information in such a way.  Specifically, we address two research questions:

\textbf{RQ1}: How good are end-to-end large language models used alone for generating hedges when fine-tuned on a peer-tutoring dialogue dataset? Are the models able to implicitly learn when and how to generate hedges? \label{rq1}

The first question may be answered by comparing the performance of various fine-tuned models. If the end-to-end models cannot learn to hedge implicitly, we might attempt to drive the models to generate the utterances by providing the correct labels. We assume that the correct labels can be provided by another module of the system, so we compare the reranking method  with the fine-tuning method, as the former is simple, powerful, and widely used for text generation. Consequently, the second question is:

\textbf{RQ2}: Can we improve these models by using a reranking approach? If so, what are the remaining errors and why do they occur? \label{rq2}


\section{Related Work}\label{related_work}

Considerably more computational methods exist to determine \textit{what} a dialogue system should say than \textit{how} to say it. However, more recently, with the increased power of end-to-end models to find information and convey it accurately, we can now turn to ensuring that the end-to-end model simultaneously also meets social goals, to increase the impact and acceptability of what is conveyed. 


\subsection{Theoretical Approaches to hedges}\label{theory_hedges}


As described above, a hedge can soften the impact of an utterance that might otherwise seem rude, such as a demand (``could you pass the salt'') or an instruction (``you might want to pour the coffee over the sink''). \citet{madaio2017impact} has attested to the frequent use of hedges in the peer-tutoring setting, and their positive impact on performance, perhaps because hedges in this context might reduce a tutee's embarrassment at not knowing the correct answer \citep{rowland2007a}. 

In linguistic terms, hedging is a rhetorical strategy that attenuates the full force of an expression \citep{fraser2010a} and for this reason, it has been covered in linguistic pragmatics and the study of politeness. Two main categories of hedges are identified in the literature: \textbf{Propositional Hedges} and \textbf{Relational Hedges} \citep{prince1982a}. Propositional Hedges (called \textbf{Approximators} by \citep{prince1982a}) refer to uncertain \citep{vincze2014uncertainty}, fuzzy \citep{lakoff1975hedges} and vague \citep{williamson2002vagueness} language use, such as ``kind of''. Relational Hedges (called \textbf{Shields} in \citet{prince1982a}) indicate that the expression is subjective or an opinion, as in ``\textit{I think} that is incorrect''. \textbf{Attribution Shields} are a subtype of relational hedges that attribute the opinion to others, such as ``everyone says you should stop smoking''. \textbf{Apologizers} \citep{raphalen-etal-2022-might} are apologies that mitigate the strength of an utterance, as in ``I'm sorry but you have to do your homework''. 

While the different types of hedges operate in different ways, they all serve the same mitigation functions in conversation. For this reason, in what follows --- a first attempt at generating hedges --- we collapse the different sub-classes and refer only to hedges and non-hedges. 

\subsection{Computational Approaches}

Some prior work has looked at the detection of conversational strategies and in particular work by Zhao and colleagues \citep{zhao2014towards, zhao2016a, zhao2016automatic}. \citet{madaio2017impact} built a classifier to detect hedging and achieved an accuracy of 81\%. Recent work by \citet{raphalen-etal-2022-might} improved the detection of different types of hedges and achieved a weighted F1 score of $0.97$.

Hedging is a particular kind of indirectness, and therefore as we look at prior work in the area, we include approaches to the generation of indirect speech. The plan-based generation of indirect speech acts has existed almost as long as dialogue systems themselves \citep{clark1979responding, brown1980characterizing, perrault1980plan}. More recently, other relevant aspects of politeness have also been addressed. For example, \citet{Porayska2004modelling} operationalized the important notion of face in politeness theory to generate polite sentences with a template pool. Although contemporary dialogue systems tend to integrate indirect speech \citep{miehle2022say, briggs2017enabling}, generating hedges with powerful language models, and particularly as a function of the social context, has not been explored. Our desire to look at the social context leads us to train on spontaneous dialogue that is substantially noisier, owing to natural conversational phenomena such as disfluency. This differs from the majority of prior work, trained on written or acted corpora \citep{li-etal-2017-dailydialog, rashkin2019towards}. 

\subsection{Generation Techniques} \label{generation_tech}

Different techniques have been used in the past to generate responses of a particular kind for dialogue systems. \citet{madaan2020politeness} used n-gram TF-IDF to identify source style words and generate target politeness style utterances by replacing these words. \citet{niu2018polite} generated politeness formulations by using reinforcement learning with a trained politeness classifier. Similar to our approach, the explicit knowledge of politeness is only given to the classifier. \citet{liu2021towards} constructed an emotional support dataset with eight different dialogue strategies and fine-tuned the pre-trained language models by connecting the label tokens to the beginning of each utterance in order to create a dialogue generator that can produce the target responses without focusing on the social context.

The reranking method is also widely used in text generation tasks. \citet{hossain2020simple} used a simple and effective pipeline where they retrieved the original texts from the database, then edited with a Transformer \citep{vaswani2017attention} model, and then reranked the text by generation scores. \citet{soni2021enhancing} first applied reranking to conversational strategy generation by controlling the level of self-disclosure in the outputs of DialoGPT \citep{zhang2019dialogpt}. The authors of LaMDA \citep{thoppilan2022lamda} used various classifiers to rerank and filter out inappropriate responses. Recently, ChatGPT \citep{chatGPT2023} used reinforcement learning with human feedback, and has shown impressive performance. 

In the articles above, most algorithms were trained on written dialogue datasets, which facilitated the task. However, our spontaneous dialogue dataset may lead the way for cutting-edge models trained on a real-world, face-to-face interactional dataset.

\section{Methodology}
\subsection{Task Description}
Let $D=\{d_1,d_2,d_3,...d_n\}$ be a set of dialogues, where each dialogue $d=\{u_1, u_2, u_3...u_m\}$ is composed of $m$ turns, where $u_i$ is a turn. Each tutor turn (and each tutee turn, although we will not examine the tutee turns further here) is labeled as hedge or non-hedge; we call $l_i$ the label of $u_i$. A fixed window size $\omega$ of the dialogue history is assigned to each utterance: $h_i=\{u_{max(1, i-\omega)}, u_{i-\omega+1},...u_{i-1}\}$. The goal of this work is to train a generator ($G$) that can produce a tutor's utterance $u'_i$ that matches a given hedge strategy (i.e., hedge or non-hedge) $l_i$, according to the dialogue history $h_i$.

\subsection{Corpus}
The dataset we used in the current work is the same as that used in our prior work \citep{raphalen-etal-2022-might, goel2019a, zhao2014towards}. 24 American teenagers aged 12 to 15, half boys and half girls, were assigned to same-gender pairs. They took turns tutoring each other in linear algebra once a week for five weeks, for a total of 60 hours of face-to-face interaction. Each interaction was composed of two tutoring periods, where the teens took turns being the tutor, with a social period at the beginning and between the two tutoring periods. For the purposes of the earlier work the corpus was  annotated for hedges, as well as the subcategories of hedges, at the clause level. For our purposes, since generation happens at the level of the turn, we merge the clauses and their labels into speaker turns and turn-level hedge labels (see Appendix \ref{clause_2_turn} for the merge strategy).

Our goal is to create a hedge generation module that can produce an appropriate hedge strategy for a tutor giving an instruction, according to what has been said before as indicated by the dialogue history. For this reason we kept all turns in the dialogue history, even though our model is trained to generate only the tutor's turns (and not those of the tutee).  There are 6562 turns in these interactions, of which 5626 contain non-hedges and 936 hedges. 


Being authentic interaction, there are disfluencies (``so just yeah just um''), repetitions (``that would be then that would be''), repairs (``oh wait, actually the x would go here''), and other spoken phenomena such as one-word clauses. These phenomena make generating hedges challenging since the language models we use are primarily trained on written dialogues, which do not contain most of these features. However, our work allows us to see how far we can go with authentic spoken data. 

\subsection{Methods}
We combine two techniques for generating the tutor's turn: \emph{Fine-tuning} an existing generation model and \emph{Re-ranking} the generated outputs to match the desired hedge strategy.

\subsubsection{Fine Tuning Method}
First, we want to evaluate how well the model performs when hedge information is implicitly taught through fine-tuning.
We fine-tuned the generation model with the training set of the peer-tutoring corpus. Each utterance $u_i = (w_1,...,w_n)$ is composed of $n$ tokens, the dialogue history $h_i$ as input to the generation model. We apply cross-entropy loss between $u_i$ and $u'_i$, where $u' \in R^{|V|}$, $V$ is the vocabulary.

\begin{equation}
    J(u_i,u'_i) = -\frac{1}{n}\sum_{j=1}^{j=|V|} u_{i,j}\log(u'_{i,j})
    \label{eq:fine-tuning}
\end{equation}
    
\subsubsection{Reranking Method}
Since a hedge classifier was developed for prior work in our lab \citep{goel2019a, raphalen-etal-2022-might}, we can use it to determine whether a generated text is a hedge or not and then inform the generator of the decision in order to regulate the output. This is known as reranking, and is what we use here as our second generation strategy.

\begin{figure*}
    \centering
    \includegraphics[width=0.8\textwidth]{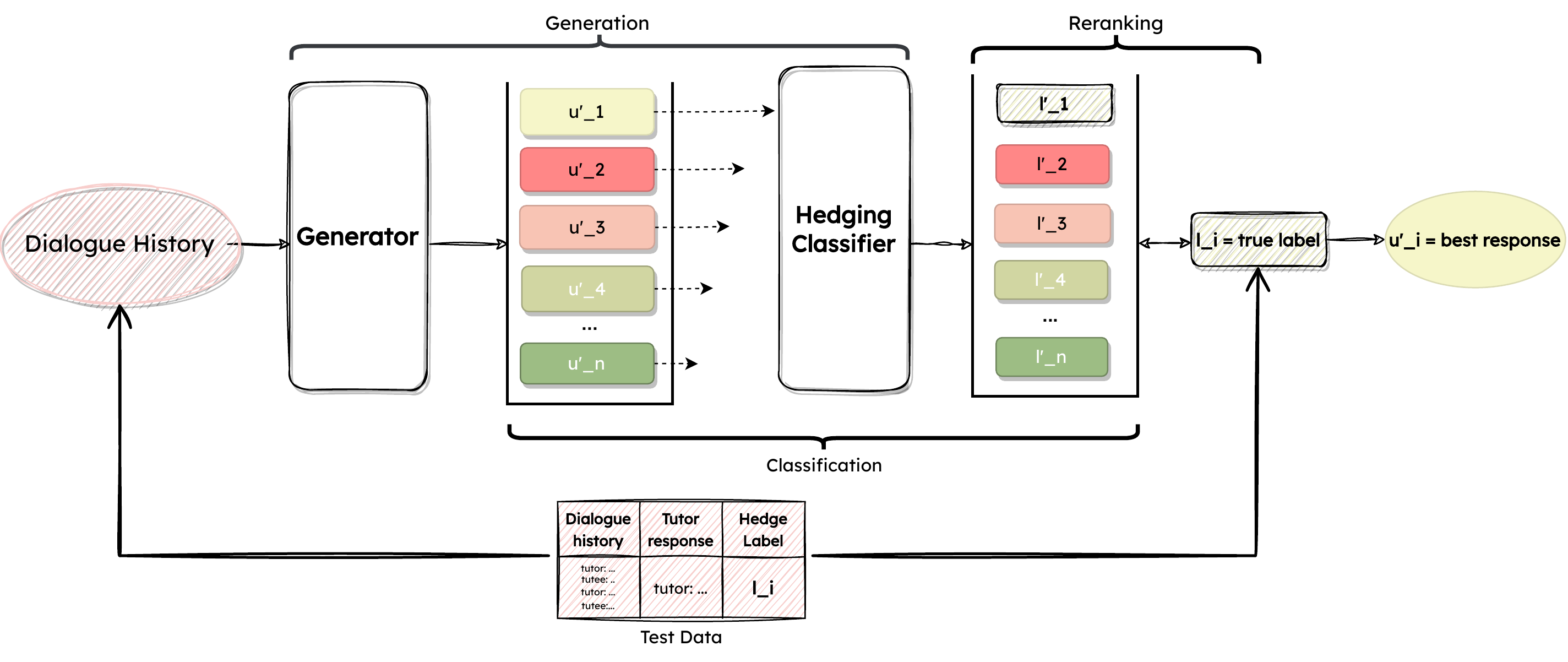}
    \caption{\footnotesize{Reranking method}}
    \label{fig:reranking}
\end{figure*}

1) We first pretrain our generator as in fine tuning. We then apply this generator to the test set to generate 50\footnote{See Appendix \ref{appendix:imp_details} for the details} candidate utterances for each dialogue history (Figure \ref{fig:reranking}). 2) These candidates are first ranked by their sentence scores (i.e., the final outputted token's log probability for each sentence). 3) We then use the hedge classifier described above to filter out the utterances that do not match the selected strategy (i.e., hedge or non-hedge). 4) We keep utterances that match the selected hedge strategy. If more than one candidate matches the strategy, we pick the first one that matches, which means the one with the highest sentence score. 5) If none of the candidates matches the selected hedge strategy, we output the one that has the highest sentence score. 



\section{Experimental Setting}

\subsection{Data Processing}
We randomly split the final dataset based on a 60:20:20 ratio. Of these, 60\% is the training set, 20\% is the validation set, and 20\% is the test set.  

Since our dataset is highly unbalanced, if we used it as is the results would be too biased towards non-hedges. In that approach the gap between the results of different models would not be clear because non-hedges are so much more frequent. For this reason, we manually balance by randomly selecting 235 non-hedge turns to balance the 235 hedges in the test set, and combine the data to form a new balanced test set. On the other hand, in order to have a large enough training set, we retain all tutor turns from the complete dataset, which therefore consists of 701 hedge turns and 4455 non-hedge turns, resulting in a dataset that is very skewed, but has more turns. 

While the complete dataset contains a relatively small number of hedge turns, we believe that preserving the natural data distribution is crucial for addressing our first research question. Underscoring the wisdom of this approach, the results we obtained on perplexity and the BARTscore (that are indicative of fluency in the generated responses, as described below) demonstrate that the models were able to generate responses with reasonable fluency and quality despite the small number of hedge turns.


\subsection{SOTA Pretrained Language Models} \label{sota}
We compare the performance of different state-of-the-art (SOTA) free open-source pretrained models as our generators: BART, DialoGPT, and BlenderBot. 
BART \citep{lewis2019bart} uses an encoder-decoder architecture, trained on books and Wikipedia data, and performs well on tasks as varied as Q\&A (SQuAD \citep{rajpurkar-etal-2016-squad}), text generation, text classification (MNLI \citep{williams-etal-2018-broad} ), and text summarization tasks (ELI5 \citep{fan-etal-2019-eli5}). It is pretrained by distorting the format of the input text in various ways, and this training helps us to visualize its possible application to noisy spontaneous spoken dialogues. DialoGPT \citep{zhang2019dialogpt} is a dialogue version of GPT-2 \citep{radford2019language}, an autoregressive language model with a multi-layer Transformer \citep{vaswani2017attention} decoder as its model architecture. It is trained on 140 million conversational exchanges extracted from Reddit comment threads. BlenderBot \citep{roller2020recipes} uses the standard Seq2Seq Transformer architecture, but incorporates a number of dialogue training sets: Empathetic Dialogue \citep{rashkin2019towards}, PersonaChat \citep{zhang-etal-2018-personalizing}, ConvAI2 \citep{dinan2020second}, and other datasets that, while largely handcrafted, focus on personality and emotions, enabling it to potentially develop some version of social skills.

\subsection{Evaluation Metrics}

To evaluate performance, we used the most widely used set of reference-based metrics for natural language generation tasks \citep{liu2021towards, ziems-etal-2022-inducing}. Since these metrics have not been used for conversational strategies, we add an unsupervised reference-free metric, the BART score \citep{yuan2021bartscore}. The BART score formulates the evaluation process as a text generation task using a pre-trained model. The score represents the probability of generating a hypothesis given a source text. The higher BART score represents better text from different perspectives (e.g., informativeness, factuality). In this paper, we denote the dialogue history as the source text and the generated utterance as the hypothesis. For comparison, we calculate the BART score between the dialogue history and the real response in the test dataset, giving a result of  $-6.44$. We also evaluated the relevance of the generated hedge strategy using an F1 score. The results using these metrics are presented in Table \ref{table:baselines}. The detailed description of the metrics used is in Appendix \ref{appendix:metrics}. 

\subsection{Human Evaluation}\label{human_evaluation}

While the metrics described above are important for comparison with the performance of other work in the field, they do not obviate the need for human annotation. We therefore asked two annotators to ignore sub-categories and annotate only hedge or non-hedge on each tutor turn of the model's output, with access to 4 prior turns of the dialogue history.  During a training phase  the annotators reached an inter-rater reliability of over $.7$ Kripendoff’s alpha \citep{krippendorff2004reliability} which indicates substantial agreement. One of the annotators then finished the remainder of the annotation. We computed the F1 scores for the label of the generated utterances with respect to the real tutor turn's  label. A higher F1 score indicates that the approach is better suited to generate the correct hedge strategy (see Table \ref{table:baselines}). We also asked the annotators to pay attention to whether the output was unnatural and to note it if so. The annotators reported no concerns with the naturalness of the generated utterances.

The concept of fluency has recently gained popularity in the dialogue community \citep{li2019incremental, see2019makes}, but the current definition of fluency varies. More fundamentally, evaluations of this kind are more applicable to written text or scripted dialogues \citep{pang2020towards, d2019automatic}. as they cannot handle disfluencies (e.g., hesitations, repetitions, false starts) of the kind that are common in spontaneous spoken dialogues, and that may serve to give the speaker time to plan the next utterance \citep{biber1999longman, thornbury2006conversation}. We therefore did not assess fluency in this work.

\section{Results}

\subsection{RQ1: How well do end-to-end models perform alone for generating hedges?} \label{RQ1}

Table \ref{table:baselines} compares the performance of the generation models. BlenderBot outperforms the other 2 models on most metrics,although with similar performance to DialoGPT, on BLEU and ROUGE-L. The discrepancy between BlenderBot and BART in each score is relatively wide. This discrepancy is most apparent on measures that compute scores based on n-gram-level overlaps (BLEU, ROUGE). To find the reason for this discrepancy, we calculate the average length of the outputs of the 3 models and observe 5.2 words for BART, 11.8 words for BlenderBot, and 14.5 words for DialoGPT, while the average length of the tutor's utterances in test data is 15.2 words.  The average length of the output of DialoGPT is therefore close to that of the test set. This further explains DialoGPT's strong performance on the BLEU and ROUGE scores. On the other hand, BART tends to generate shorter turns, consequently demonstrating lower scores on metrics that require the calculation of repetition grams to yield scores. Note that in similar tasks, the best model  was Blenderbot with a BLEU 2 score of 6.21, in the case of emotional support conversational strategy generation \citep{liu2021towards}, while DialoGPT reached 5.52. The best score in the positive text reframing task, meanwhile, was 11.0 for BLEU 1 \citep{ziems-etal-2022-inducing}, while BART reached 10.1 and GPT-2 reached 4.2.

Table \ref{table:ppl} shows that BART has the lowest perplexity score, indicating that BART is more adaptive to our dataset compared to the other two models. This may be due to its pre-training approaches (see Section \ref{sota}) that corrupt input texts with an arbitrary noising function. These approaches enable more accurate predictions in our noisy real-world dataset.


\begin{table}[!ht]
    \centering
    \resizebox{0.6\linewidth}{!}{
    \begin{tabular}{lll}
    \hline
        BART & BlenderBot & DialoGPT \\ \hline
        34.9 & 69.3 & 72.4 \\ \hline
    \end{tabular}
    }
    \caption{\footnotesize{Language Model (LM) Perplexity (the lower is the better}}
    \label{table:ppl}
\end{table}

In response to our first research question, then, the performance of all three models was comparable but very limited. This suggests that the fine-tuning approach does not allow language models to learn hedge knowledge implicitly.


\begin{table}[!ht]
    \centering
    \resizebox{\linewidth}{!}{
    \begin{tabular}{p{1.1cm}|p{1.1cm}p{1.1cm}p{1.1cm}|p{1.1cm}p{1.1cm}p{1.1cm}}
    \hline
        \diagbox[width=1.5cm]{\scriptsize{Metrics}}{\scriptsize{Models}} & \footnotesize{BlenderBot} & \footnotesize{DialoGPT} & \footnotesize{BART} & \scriptsize{R\_BlenderBot} & \scriptsize{R\_DialoGPT} & \scriptsize{R\_BART} \\ 
        \hline
        \footnotesize{BLEU\_1} & 11.2 & \textbf{11.4} & 2.7 & \textbf{12.3} & 10.9\oneS & 6.0\oneS \\ 
        
        \footnotesize{BLEU\_2} & \textbf{5.8} & 4.7 & 1.5 & \textbf{6.2} & 3.9\oneS & 3.1\oneS \\ 
        
        \footnotesize{ROUGE-L} & 8.6 & \textbf{9.1} & 8.1 & \textbf{11.0} & 8.4 & 9.7 \\ 
        
        \footnotesize{CHRF} & \textbf{17.6} & 17.0 & 9.3 & \textbf{17.6}\oneS & 17.5\oneS & 12.2\oneS \\ 
        
        \scriptsize{BARTScore} & \textbf{-3.92} & -5.62  & -4.33 & \textbf{-3.98\oneS} & -4.79 & -4.24 \\ 
        \scriptsize{BERTScore} & \textbf{39.9} & 38.3 & 38.5 & \textbf{40.5} & 37.5 & 39.4 \\
        \hline
        \footnotesize{F1Score} \scriptsize{(human evaluation)} & \textbf{0.54} & 0.41 & 0.44 & 0.84 & 0.64 & \textbf{0.85} \\ 
        \hline
    \end{tabular}
        }
    \caption{\footnotesize
    {Results of the fine-tuned models and reranking method applied to the fine-tuned models.    \oneS  means this result is significantly different from the fine-tuning method ($p < .05$)}
    }
    \label{table:baselines}
\end{table}

We therefore next turn to an approach that may improve performance by screening utterances with a given label. 

\subsection{RQ2: Does reranking  improve hedge generation?}\label{section5.2}


Table \ref{table:baselines} shows the performance of each model for the reranking method. BlenderBot once again performs well on all metrics and has a virtually identical F1 score to BART. Additionally, we find some interesting similarities among models: 1) BlenderBot and DialoGPT outperform BART in both the fine-tuning and the reranking methods (Table \ref{table:baselines}) with respect to reference-based metrics such as BLEU, ROUGE-L, etc., and 2) DialoGPT still underperforms the other two models in terms of F1 score, and in the reranking condition the gap widens.

This result could suggest that 1) the pretraining of the models (i.e., DialoGPT, BlenderBot) on dialogue datasets may help to generate longer utterances, and therefore to improve the reference-based metrics performance, and 2) the autoregressive model (e.g., DialoGPT) may not be suitable for the generation of social dialogue such as hedges.


\subsection{Comparing Fine-tuning and Reranking}

To summarize results on the fine-tuning versus re-ranking approaches we observe that: 1) With the help of a hedge classifier, the reranking approach can do a good job at generating hedges, 2) BlenderBot is better suited to the task of generating long utterances, as described in Section \ref{RQ1}. This could be because BlenderBot is pretrained with various social dialogue datasets, giving it a certain ability to generate the social aspects of dialogue. 

Table \ref{table:baselines} shows that models deployed with the reranking method have relatively higher or comparable Bart scores, but greatly improved performance on the F1 score (from $.54$ to $.85$). This result, too, underscores the advantages of the reranking method.

\subsection{Error Analysis}\label{error_analysis}

While BlenderBot showed strong performance when using reranking, a certain number of generated utterances still did not match the real tutor labels. When a matching utterance type cannot be found in a limited pool of candidates, we could have chosen to increase the candidate pool to promote the probability of selecting a match.  However, in this early effort to generate hedges, we want to ensure sufficient quality in the generated output but also explore the limitations of current language models for generating socially relevant phenomena on the basis of a spontaneous spoken interaction dataset.

We can learn about the limitations of these models by examining places where the system did not generate the desired strategy (that is, generated a hedge when the real tutor did not or vice versa). We first divide these strategy mismatches into \textit{over-generation errors}, where the generator generates a hedge where it should not and \textit{under-generation errors} when it does not generate a hedge but should. Among the 1395 annotated turns outputted by the 3 generators, there are 13.3\% of \textit{over-generation errors} and 86.7\% \textit{under-generation errors}. These errors are particularly interesting in the context of reranking, as it relied strongly on the hedge classifier. The hedge classifier selected the most suitable utterances, and yet the model still produced the wrong strategy - or at the very least mismatches with the strategy of the real tutor. 

\begin{figure*}
    \centering
    \includegraphics[width=0.7\textwidth]{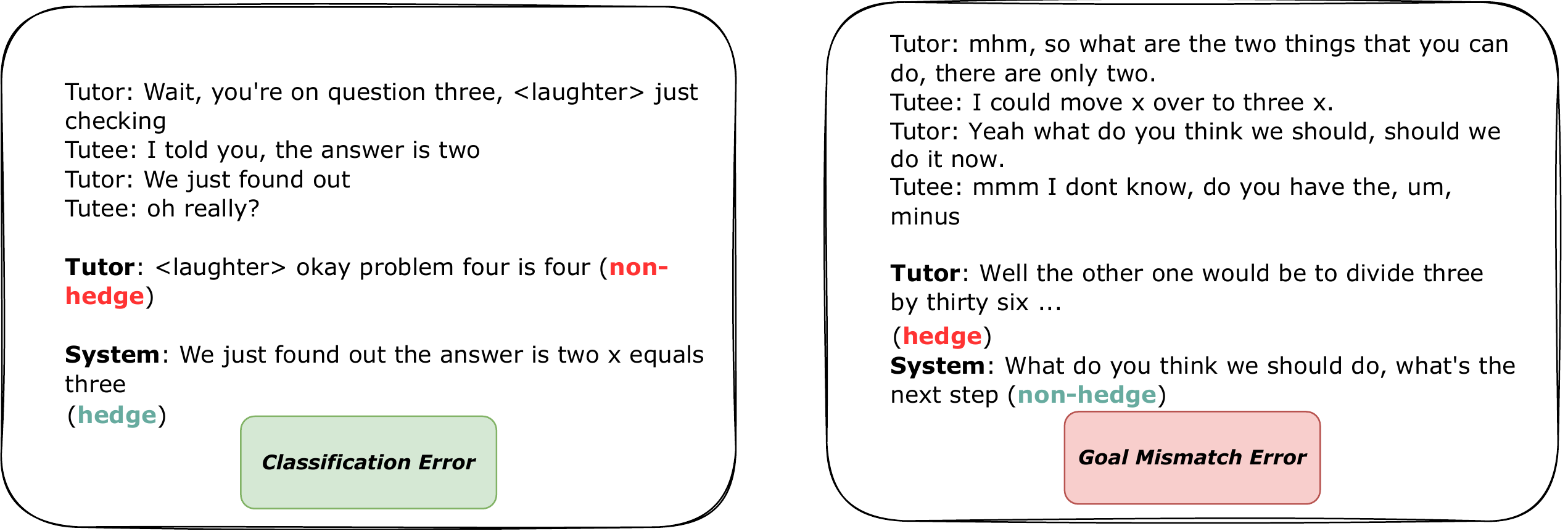}
    \caption{\footnotesize{Strategy Mismatch Errors for Reranking Method}}
    \label{fig:type_errors_reranking}
\end{figure*}

Therefore, we analyze the generated utterances corresponding to these two types of errors and identify two potential causes.

First, there are still some places where the model generates a hedge where it should generate a non-hedge. As we mentioned in Section \ref{human_evaluation}, we invited humans to annotate the models' outputs in terms of hedge labels. We compare the human-annotations of the model output (where they labeled the output as hedge or non-hedge) with the output of the BERT-based classifier on the same generated utterances to calculate the F1 score. We find that there is a difference of about 9 points between the F1 score for human annotation (85\%) shown in Table \ref{table:baselines}, and the F1 score for the same BERT-based hedge classifier (94\%) reported in \citet{raphalen-etal-2022-might}. We assume that the classifier we used may have misclassified some generated utterances and we therefore label them as \textbf{Classification Errors}. This category accounts for 92.5\% of \textit{over-generation errors}, and 15.3\% of \textit{under-generation errors}.
    
Second, the basic functionality of an end-to-end language model of this kind is to produce the most coherent next utterance based on the dialogue history. This may result in the language model privileging coherence of content over style of delivery. That is, the model may not be able to find an appropriate strategy match among the coherent candidates, even when the candidate pool size is 50. We label this a \textbf{Goal Mismatch} as the propositional or content coherence goals of the system may be trumping the social goals, We found 84.7\% in \textit{under-generation errors} and 7.5\% in \textit{over-generation errors}. 18\% of the cases where the pool did not include the right strategy.

An example of each type of error is given in Figure \ref{fig:type_errors_reranking}. The first example belongs to the \textbf{Classification Error} type, where the classifier misclassified the system response (i.e. ``We just found that the answer is two x equals three'') as a hedge. In the second example, the tutor is trying to help the tutee to approach the answer step by step, but the tutee cannot come up with a worked idea. Here it is clear that the tutee is flailing and it is therefore probably not advisable to increase the student's stress with a volley of questions that the tutee can clearly not answer. The tutor thus uses a hedge as a response. Conversely, the generator produces a question. The generated utterance is ``What do you think we should do, what's the next step''.  This example corresponds to our \textbf{Goal Mismatch Error}. It shows that the generator may not understand the social context, but is looking for a coherent response.  

The \textbf{Goal Mismatch Error} is perhaps the most interesting of the errors, and thus to verify our hypothesis --- that the coherence goals of the models may impede the social goals --- we looked into the nature of the relationship between rapport (between tutor and tutee) and the generation of hedges. As described above, \citet{madaio2017impact} found that hedges are generated when rapport is low. Since our corpus contained rapport annotations for every 30 seconds of the interaction, we looked at the rapport level in play when the model over-generated and under-generated hedges. Since rapport is annotated from 1 to 7 in the dataset, for convenience, we divided it into 3 levels: high (5-7), medium (3-5), and low rapport (1-3), as shown in Table \ref{table:error_rapport}.

\begin{table}[!ht]
    \centering
    \resizebox{0.7\linewidth}{!}{
    \begin{tabular}{l|lll}
    \toprule
        \diagbox[]{Type}{Rapport} & \textbf{High}  & \textbf{Medium}  & \textbf{Low} \\ \midrule
        \textbf{Over-generation} & 0 & 3 & 0 \\
        \textbf{Under-generation} & 13 & 130 & 75 \\ \bottomrule
    \end{tabular}
    }
    \caption{\footnotesize{Goal Mismatch Errors Distribution}}
    \label{table:error_rapport}
\end{table}

As only 3 errors appear in the category of \textit{over-generation error}, we cannot obtain a meaningful conclusion due to size. However, the generators generate fewer hedges when rapport is low, an \textit{under-generation error}, in contradiction to studies showing that speakers are more careful about threatening the face of (or embarrassing) their interlocutors when the social bond between them is weak \citep{madaio2017impact}. We believe that this is because more hedges are found in low rapport interaction. Therefore, we count the hedge distribution of the low and high rapport interaction in the test dataset. 264 hedges are found in low rapport interaction, and 42 in high rapport interaction. This distribution corresponds to the fact that a hedge is most likely to happen in low rapport interactions. The under-generation errors are the cases where there should be hedges but non-hedges were generated. In the test dataset, more hedges occur in low rapport, and the generator under-generates more in low rapport, because there are more hedges that should be generated in low rapport. So, the generators make more errors in low rapport interaction due to an imbalance in hedge distribution between low and high rapport interaction.


Goal Mismatch error directly addresses our primary question \ref{rq1}: How effectively do end-to-end models perform when generating hedges on their own? Due to this fundamental discrepancy between competing goals, end-to-end language models are unable to inherently learn and discern when to apply hedges appropriately. 

\subsubsection{Lexical Diversity of the Generated Output}

As we have seen, LLMs can generate a hedge or non-hedge with the help of the reranking method. However, do language models spontaneously use different types of hedges in a human-like way? To investigate this question, we applied the rule-based hedge classifier from \citep{raphalen-etal-2022-might} to automatically annotate the utterances generated by models in subcategories of hedges (as defined in Section \ref{theory_hedges}), and we compare the models' and humans' distributions of different hedge strategies. The rule-based classifier used linguistic patterns to identify each hedge subcategory. We have preferred here to use the rule-based classifier rather than the machine learning classifiers  to avoid the dependence on and bias of probabilistic learning-based classifiers. Indeed, learning-based classifiers may be biased towards predicting the categories that are the most frequent in the dataset. Furthermore, the rule-based classifier reaches a 94.7 F1 score \citep{raphalen-etal-2022-might}, which is comparable to the best performance (96.7 F1 score) using the Light Gradient-Boosting Machine (LGBM) \citep{ke2017lightgbm} classifier. 


The above results show that the model can spontaneously learn to use different types of hedges. Indeed, the models are capable of carrying out linguistic diversity on hedges based on learning from real human dialogues.

\section{Conclusion and Future Work}

In this paper, we have shown that the reranking method helps LLMs to generate hedges --- an important social conversational strategy that can avoid face threats towards an interlocutor by attenuating an impact of an expression. We find that an implicit fine-tuning approach (i.e., without any supervision by a hedge label) is not sufficient for generating hedges, but a reranking method significantly improves performance in generating hedges, with a final F1 score of $.85$ for the BART model and $.84$ for the BlenderBot model. We also performed an error analysis on the generated results and found that two types of errors occur in the reranking method: \textbf{Classification}, and \textbf{Goal Mismatch}. The vast majority of errors fall into the category of Goal Mismatch, indicating an important conflict between contemporary language models' primary goal of ensuring coherence and the social goal of managing face, which is indispensable for human conversation. While we were able to generate hedges, we were not able to necessarily generate them where they were needed most. That is, conversational strategies are adaptive in the sense that they respond to conversational strategies uttered by the previous speaker \citep{zhao2014towards}. We conclude that, going forward, we will need a way of adding an underlying representation of the social state of the dialogue to improve dialogue generation. 

In this paper we addressed the question of how to generate hedges, but when to generate hedges remains an important and unexplored question. In future work, we may first explore the temporal relationships between the hedge and other conversational information (e.g., other conversational strategies, level of rapport) by sequential rule mining techniques, then apply RL-based methods to investigate in a more detailed manner the optimal way to predict where hedges should occur.  In this context, we note that ChatGPT can generate a hedge when requested explicitly to do so, but does not generate hedges of its own volition (so to speak), for example, when face-threatening acts such as instruction are engaged in.

We began this paper by describing the need for hedges in instructional dialogues such as those engaged in by intelligent tutoring systems. The current dataset consists of authentic real-world tutoring sessions, but as carried out by untrained teenagers. We note that peer tutoring is a powerful method of teaching, used in classrooms around the world, and previous work shows that when untrained peer tutors use hedges, their tutees attempt more problems and solve more problems correctly \citep{madaio2017impact}. However, they are inexperienced and so in future work it will be important to investigate the interaction between trained tutors and tutee as well, for instance, by using the Teacher-Student Chatroom Corpus \citep{caines2020teacher}.  We believe that the methods and results from the current work will facilitate the investigation of expert tutors in future research.

\section*{Broader Impact}
Since the 1990s, research has shown the the importance of intelligent tutoring systems as effective learning environment,s and supports for classroom learning \citep{anderson1995cognitive}. Peer tutoring plays a powerful role as well, as peer tutors can motivate learners to try harder, as well as helping them to succeed, and it is particularly effective for low-achieving learners \citep{cassell2022socially}. But virtual peer tutors have not yet achieved their potential, in part because of the difficulty of generating the social infrastructure of peer learning as well as the content of the matter being tutored. This paper, whose data comes from a corpus of peer tutoring dialogues, should therefore be seen as a step in the right direction. 

\section*{Acknowledgments}
We thank the anonymous reviewers for their helpful
feedback. We express sincere gratitude to the members of the ArticuLab at Inria Paris for their invaluable assistance in the successful completion of this research, and to the members of the ArticuLab at Carnegie Mellon Pittsburgh for answering our questions about their prior work. This study received support from the French government, administered by the Agence Nationale de la Recherche, as part of the "Investissements d'avenir" program, with reference to ANR-19-P3IA-0001 (PRAIRIE 3IA Institute).

\section*{Limitations}

Several limitations apply to the current study. While research shows that multimodal signals play an important role in conversational strategies \citep{zhao2016a}, we did not take them into account. It is an open question as to how to render large language models capable of generating multimodal behaviors. A second limitation concerns the recent arrival on the scence of ChatGPT, that has shown impressive performance. However the models are not free, and therefore were not included. As noted above, another important limitation is the untrained status of the tutors in our corpus, who are teenagers, and not trained tutors. Their use of hedges, therefore, comes from their knowledge of everyday social interaction, and not from expertise in teaching. In looking at the data, we find a few places where, as instructors ourselves, we believe that a hedge is important, even though the real (teenage) tutor did not use one. 

The last limitation is that, while we focused only on generating hedge or non-hedge, there are actually  3 different kinds of hedges, that function differently. We hope to extend this work and take advantage of a text style transfer technique to generate more kinds of hedges in future work.  

\section*{Ethical Statement}
The corpus used here comes from earlier work by the last author and her colleagues, and was used in accordance with the original experimenters' Institutional Review Board (IRB). Those experimenters also anonymised the data, removing any identifying information. A pixelated example of the video data is available at \url{github.com/neuromaancer/hedge_generation}. To counteract potential gender bias concerning the use of hedges in peer tutoring, the data was collected from equal number of boys and girls. In text generation tasks, it is important to be aware of the potential risk of generating inappropriate content. We believe that, in fact, hedges used by tutors are perhaps the least likely conversational strategy to be inappropriate, as they are the most polite and ``delicate'' conversational moves. But, more generally, considerable additional work would be needed to filter out all inappropriate language for safe tutoring systems that engage in social and task interaction. 

\bibliography{refs}
\bibliographystyle{acl_natbib}

\appendix

\section{Clauses to Turns} \label{clause_2_turn}

In our task formulation, a dialogue is composed of tutor-tutee turns. However, in the corpus considered for this study, the available annotations are at the clause\footnote{A clause consists of a subject and a verb and expresses a complete thought \citep{berry2010identifying}.} level. The choice of annotation unit was made because the annotation in hedges was part of a larger annotation campaign dedicated to the annotation of various conversational strategies (e.g., praise) at the clause level. This corpus contains 23 156 clauses, of which 21 192 contain non-hedges and 1 964 hedges. In order to obtain annotations at a turn level, we apply the simplest way to merge the hedge labels. If one or multiple clauses of one turn are annotated as hedges, this turn is labeled as a hedge.



\section{Metrics} \label{appendix:metrics}

\textbf{BLEU} \citep{papineni2002bleu} calculates the word overlaps between reference and candidate utterances in n-grams (n=1, 2, 3).  We do not assume that higher BLEU scores are equivalent to better task completion. Instead, BLEU is used to indicate that the generated utterances retain certain desired keywords. 

\textbf{ROUGE-L} \citep{lin2004rouge} supplements BLEU by computing the longest common subsequence of generated utterances and references, allowing it to compute overlap measures in longer utterances. To avoid generated utterances that are too long for the BLEU score, we use Rouge-L as a complementary metric.

\textbf{CHRF} \citep{popovic2015chrf} is comparable to BLEU; however, while BLEU is word-level, CHRF is character-level, based on character n-gram computation. Our transcribed dataset also shows some disfluencies and repetitions represented by individual characters. Therefore, we expect this metric to result in character-level overlap scores.

\textbf{BERTScore} \citep{zhang2019bertscore} embeds the generated utterances and the reference with word vectors using the BERT model and computes pairwise cosine similarity for each generated word vector and each word in the reference, then the recall of the generated sequences is calculated. BERTScore is distinct from the previous two metrics in that it computes similarity across semantic space and has been shown to have a strong correlation with human judgment at the segment level.

\textbf{BARTScore} \citep{yuan2021bartscore} formulates the text generation evaluation as a text generation task from pretrained language models in an unsupervised fashion. When the generated text is better, the training model will get a higher score by converting the generated text to reference or source text. BART score can be applied to different evaluations (e.g., informativeness, coherence, and factuality).

\textbf{Perplexity} \citep{chen1998evaluation} calculates  language model perplexity. Perplexity quantifies the level of uncertainty when an LM generates a new token.

\section{Implementation Details}\label{appendix:imp_details}
The implementation of all models was based on the Transformer library\footnote{\url{github.com/huggingface/transformers}}, in addition, the Pytorch-Lightning\footnote{\url{github.com/Lightning-AI/lightning}} library was used for training control. We apply AdamW \citep{loshchilov2018decoupled} as our optimizer with a learning rate $10e^-5$. All the models are trained with 10 epochs but with an Early-stopping mechanism on validation loss, which means when the validation loss remains for 2 epochs, the training will stop to prevent overfitting.
We use the base version of the BART model, the small version of BlenderBot, and also the small version of DialoGPT. For the reranking method, we use beam search as our decoding strategy. To prevent repetition, we allow the 2 grams to occur only once, and the repetition penalty = 1.2 is also applied. All models were fine-tuned on an Nvidia Quadro RTX 8000 GPU. A complete configuration of the hyperparameters used for each model is reported in the GitHub repository with the code of the
paper: \url{github.com/neuromaancer/hedge_generation}.

Moreover, we apply beam search for the decoding strategy, as it reduces the risk of missing hidden high-probability word sequences by retaining the $n$ most likely words in each generation output and ultimately selecting the utterances with the highest overall probability. To avoid repeating the same subsequences, we apply a penalty to the repeated 2-gram unit. In terms of the size of the candidate pool, logically, the more candidates generated, the more chances that one of them is the right hedge strategy (i.e., hedge or non-hedge), so we fix our candidate pool size to 50, as a compromise between the likelihood of obtaining a hedge and the speed of generation.

\section{Figures} \label{figure:li_diver}

\begin{figure}
    \centering
    \includegraphics[width=\linewidth]{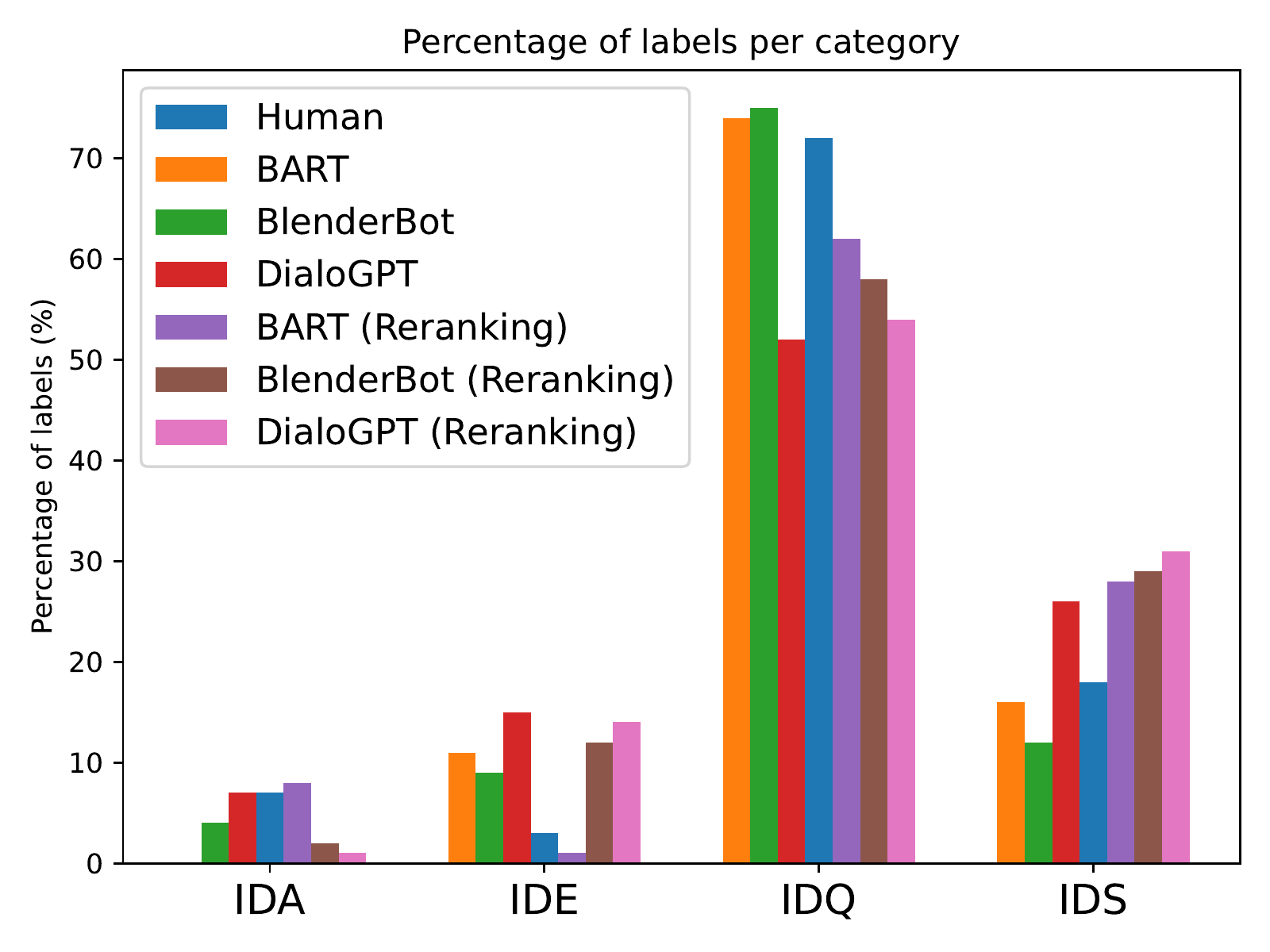}
    \caption{Hedge subcategories distribution in models' outputs compared with human. IDA: Apologizer; IDE: Extender; IDQ: Propositional hedges; IDS: Subjectivizer (as defined in Section \ref{theory_hedges})}
    \label{fig:label_per}
\end{figure}

Figure \ref{fig:label_per}: Hedge subcategories distribution in models' outputs compared with human.

\end{document}